# Rotated Object Detection via Scale-invariant Mahalanobis Distance in Aerial Images

Siyang Wen, Wei Guo, Yi Liu and Ruijie Wu

*Abstract*—Rotated object detection in aerial images is a meaningful yet challenging task as objects are densely arranged and have arbitrary orientations. The eight-parameter (coordinates of box vectors) methods in rotated object detection usually use $l_n$-norm losses (L1 loss, L2 loss, and smooth L1 loss) as loss functions. As $l_n$-norm losses are mainly based on non-scale-invariant Minkowski distance, using $l_n$-norm losses will lead to inconsistency with the detection metric rotational Intersection-over-Union (IoU) and training instability. To address the problems, we use Mahalanobis distance to calculate loss between the predicted and the target box vertices' vectors, proposing a new loss function called Mahalanobis Distance Loss (MDL) for eight-parameter rotated object detection. As Mahalanobis distance is scale-invariant, MDL is more consistent with detection metric and more stable during training than $l_n$-norm losses. To alleviate the problem of boundary discontinuity like all other eight-parameter methods, we further take the minimum loss value to make MDL continuous at boundary cases. We achieve state-of-art performance on DOTA-v1.0 with the proposed method MDL. Furthermore, compared to the experiment that uses smooth L1 loss, we find that MDL performs better in rotated object detection.

## I. Introduction

OBJECT detection in remote sensing has a wide range of applications in city planning, disaster rescue, and military filed, which has been developing rapidly nowadays [1], [2]. Different from detecting horizontal objects in general images, object detection in aerial images focuses more on objects' orientations as objects are densely arranged and have arbitrary orientations. Therefore, rotated object detection now is applied to aerial images for high-precision detection.

Rotated object detection originates from horizontal object detection and uses the same frameworks but requires some changes due to the extra property orientation. Horizontal object detection generally uses the coordinates of center, height and width to represent horizontal boxes, which can easily get the Intersection-over-Union (IoU) between the ground truth and the predicted box. As the IoU is the way of evaluating predicted results and is differentiable, IoU loss [3] (calculated by 1-IoU) and its improved methods (e.g., GIoU loss [4] and DIoU loss [5]) are widely used as loss functions. However, the same idea doesn't work in rotated object detection as the rotational IoU (aka., SkewIoU) is much more complex and is not differentiable. Therefore, $l_n$-norm losses (L1 loss, L2 loss and smooth L1 loss [6]) mainly based on Minkowski distance are frequently used when regressing rotated box parameters.

There are usually two ways to describe the oriented bounding box (OBB) of a target object: (i) the coordinates of center, width, height and angle of the box (five parameters) (ii) the coordinates of box vectors (eight parameters) [7]. When using $l_n$-norm losses for the regression of five parameters, prediction accuracy may get hurt as the orientation is in the measurement system different from others and slight deviation will cause a drastic change in SkewIoU. As for eight-parameter methods, although the coordinates of box vectors have the same unit, it can still damage prediction performance to some extent as $l_n$-norm losses are not scale-invariant.

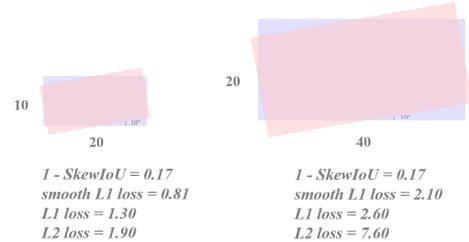

Fig. 1. Changes of 1-SkewIoU and ln-norm losses when double the scale.

Fig. 1 shows that the values of 1-SkewIoU (same idea as IoU loss) keep invariant while $l_n$-norm losses change significantly when the heights and widths of the OBBs are doubled. It reveals that the scale of the OBB has a great influence on $l_n$-norm losses, which exacerbates the inconsistency with the detection metric SkewIoU. Moreover, it can also lead to training instability as loss fluctuates when the scale varies greatly. As it it very common for aerial images to have targets with different scales, the impact of these problems cannot be ignored.

In order to reduce the impact of the above issues, in this paper we use Mahalanobis distance [8] to calculate loss between the predicted and ground truth box vertices' vectors. Different from Minkowski distance, Mahalanobis distance is scale-invariant as it takes into account the correlations of the data. Therefore, our

This work was supported by the National Key Technologies Research and Development Program of China under Grant 2016YFC0802500. *(Corresponding author: Wei Guo).*

Siyang Wen is with the State Key Laboratory of Information Engineering in Surveying, Mapping and Remote Sensing, Wuhan University, China. (e-mail: siyangwen@whu.edu.cn).

Wei Guo is with the State Key Laboratory of Information Engineering in Surveying, Mapping and Remote Sensing, Wuhan University, China and the Nanjing Beidou Innovation and Applied Technology Research Institute Co., Ltd, China. (e-mail: guowei-lmars@whu.edu.cn).

Yi Liu is with the School of Geodesy and Geomatics, Wuhan University, China. (e-mail: yliu@sgg.whu.edu.cn).

Ruijie Wu is with the State Key Laboratory of Information Engineering in Surveying, Mapping and Remote Sensing, Wuhan University, China. (e-mail: jerrywu@whu.edu.cn).

proposed loss function Mahalanobis Distance Loss (MDL) is more consistent with the detection metric SkewIoU and more stable during training. However, due to angle periodicity, MDL still has the boundary discontinuity problem like all other eight-parameter methods, which can cause loss upsurge when two close boxes are on both sides of an axis [7]. To deal with the issue, we take the minimum loss value [7] to make MDL continuous at boundary cases. The effectiveness of MDL is confirmed by experiments implemented on DOTA-v1.0 [9].

In summary, the contributions of this paper are as below:
1) We introduce a new loss function called Mahalanobis Distance Loss (MDL) for eight-parameter rotated object detection.
2) MDL can promote the consistency with detection metric and make training more stable than $l_n$-norm losses as Mahalanobis distance is scale-invariant.
3) We achieve state-of-art performance on DOTA-v1.0 with MDL. Besides, the experiments using MDL and the widely used smooth L1 loss under the same condition show that MDL is more effective and more stable during training in rotated object detection.

## II. PROPOSED APPROACH

In this section, we first introduce our new loss function called Mahalanobis Distance Loss (MDL) for eight-parameter rotated object detection, then analyze its advantages in rotated object detection. Finally, we demonstrate the overall loss function design with our architecture.

### A. Mahalanobis Distance Loss for Oriented Bounding Box

Inspired by the fact that $l_n$-norm losses are mainly based on Minkowski distance, we propose a new loss function MDL based on Mahalanobis distance. Mahalanobis distance, which is scale-invariant, can measure the standard deviations from vector $\boldsymbol{m}$ to $\boldsymbol{n}$ that follow the same distribution. The Mahalanobis distance between $\boldsymbol{m}$ and $\boldsymbol{n}$ is defined as:

$$MD(\boldsymbol{m},\boldsymbol{n}) = \sqrt{(\boldsymbol{m}-\boldsymbol{n})^T \Sigma^{-1}(\boldsymbol{m}-\boldsymbol{n})} \quad (1)$$

where $\Sigma$ refers to the covariance matrix of the distribution.

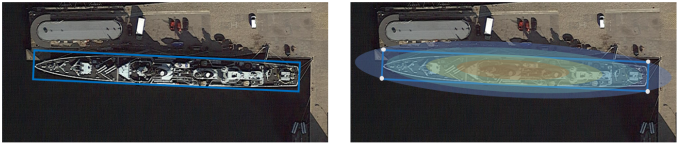

Fig. 2. Viewing OBB as a two-dimensional distribution which consists of four vertices of the box.

An oriented box can be viewed as a two-dimensional distribution which consists of four vertices of the box, as illustrated in Fig. 2. Suppose the vertices' vectors of the oriented box $B(\boldsymbol{a},\boldsymbol{b},\boldsymbol{c},\boldsymbol{d})$ are $\boldsymbol{a}=(x_a,y_a)$, $\boldsymbol{b}=(x_b,y_b)$, $\boldsymbol{c}=(x_c,y_c)$ and $\boldsymbol{d}=(x_d,y_d)$, then the covariance matrix $\Sigma$ of the distribution can be calculated as:

$$\Sigma = \begin{pmatrix} \frac{\sum_{i=a}^{d}(x_i-\bar{x})(x_i-\bar{x})}{N-1} & \frac{\sum_{i=a}^{d}(x_i-\bar{x})(y_i-\bar{y})}{N-1} \\ \frac{\sum_{i=a}^{d}(y_i-\bar{y})(x_i-\bar{x})}{N-1} & \frac{\sum_{i=a}^{d}(y_i-\bar{y})(y_i-\bar{y})}{N-1} \end{pmatrix} \quad (2)$$

where $N$ refers to the number of points ($N=4$ here), and $\bar{x}$ and $\bar{y}$ refer to the average value of vectors' x values and y values respectively. Note that $\Sigma$ does not change regardless of the starting point of the vectors.

Thus, MDL between the predicted oriented box $B(\boldsymbol{a},\boldsymbol{b},\boldsymbol{c},\boldsymbol{d})$ and the target oriented box $B^*(\boldsymbol{a}^*,\boldsymbol{b}^*,\boldsymbol{c}^*,\boldsymbol{d}^*)$ can be expressed as:

$$MDL(B,B^*) = \frac{1}{N}\sum_{i=a}^{d}MD(i,i^*)$$
$$= \frac{1}{N}\sum_{i=a}^{d}\sqrt{\left((x_i,y_i)-(x_i^*,y_i^*)\right)^T \Sigma^{-1}\left((x_i,y_i)-(x_i^*,y_i^*)\right)} \quad (3)$$

where $N=4$ as we use the mean value of four vertices' Mahalanobis distance for stable training, $MD(i,i^*)$ is according to Eq. 1, and $\Sigma$ refers the covariance matrix calculated by Eq. 2. Note that $\Sigma$ can be calculated by the predicted box vectors $\boldsymbol{a},\boldsymbol{b},\boldsymbol{c},\boldsymbol{d}$ or the ground truth box vectors $\boldsymbol{a}^*,\boldsymbol{b}^*,\boldsymbol{c}^*,\boldsymbol{d}^*$.

### B. Analysis of Mahalanobis Distance Loss

With using the covariance matrix $\Sigma$, Mahalanobis distance between two points is independent of the measurement units of the original data, which thus is scale-invariant [8]. Derived from Mahalanobis distance, MDL is also scale-invariant. Fig. 3 plots different loss curves when only scale is changing. We can find that the curve of MDL keeps invariant as same as 1-SkewIoU while L1 loss and smooth L1 loss get larger and larger with the scale, which indicates that MDL is more consistent with detection metric and more stable during training at such scale cases than $l_n$-norm losses.

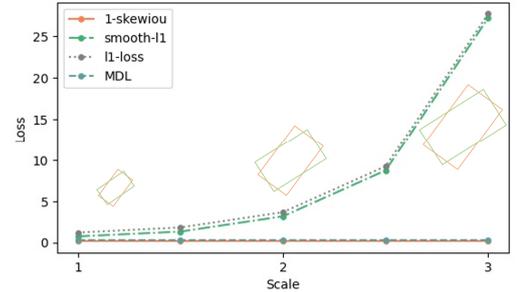

Fig. 3. Different loss curves at different scales.

Furthermore, we also compare MDL with $l_n$-norm losses at different angle, center shifting and aspect ratio cases by changing only one factor. As shown in Fig. 4, the trends of MDL loss curves are more consistent with 1-SkewIoU and more stable in all cases while the curves of L1 loss and smooth L1 loss are more steep than 1-SkewIoU.

Overall, the advantages of using MDL in rotated object detection can be summarized as:
1) MDL is more consistent with the detection metric SkewIoU than $l_n$-norm losses at different scale, angle, center shifting, and aspect ratio cases.
2) MDL can get loss values that don't fluctuate much due to its property of scale invariance, which makes training more stable than $l_n$-norm losses.

### C. Overall Loss Function Design

As anchor-based methods suffer from hyperparameters for setting anchor boxes [10], we choose the anchor-free single-stage model CenterNet [11] as the baseline. With the architect-ture shown in Fig. 5, our loss function can be divided into three parts: heatmap, box vertices and offset.

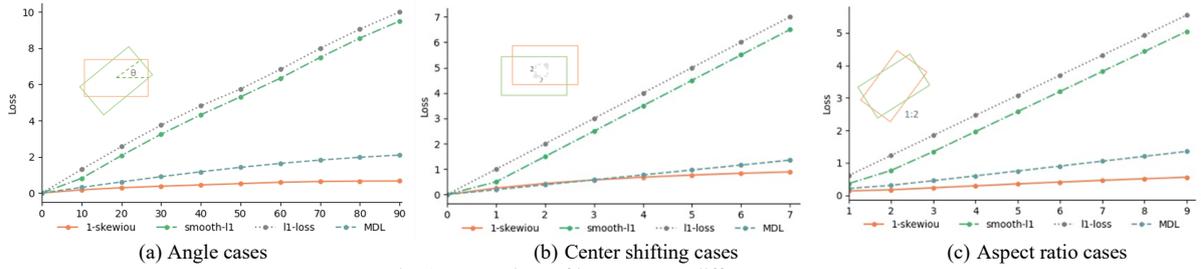

Fig. 4. Comparison of loss curves at different cases.

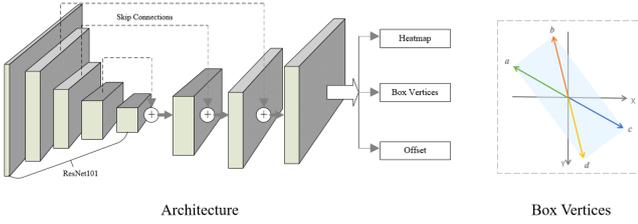

Fig. 5. The architecture and the box description of our method. We use ResNet101 as the backbone and a three-layer U-shaped structure to concatenate layers. Three parts: Heatmap P, box vertices B and offset O are the final outputs. Box vertices' vectors $a$, $b$, $c$, and $d$ are used to represent the OBB.

**Heatmap.** Heatmap is used to demonstrate where the target's center point is in CenterNet. However, it's unable to directly penalize points that are near the target center point and get large IoU between predicted box and target box. Therefore, Gaussian kernel $\exp\left(-\frac{(x-\tilde{p}_x)^2+(y-\tilde{p}_y)^2}{2\sigma_p^2}\right)$ is used to generate the ground truth and the deviation σ is adapted to the target object size. As we only view center points as the positive samples while other points are negative samples in training, we use an improved Focal loss to balance positive and negative samples as in [12]:

$$L_h = -\sum_i \begin{cases} (1-\hat{Y}_i)^\alpha \log(\hat{Y}_i) & \text{if } Y_i = 1 \\ (1-Y_i)^\beta (\hat{Y}_i)^\alpha \log(1-\hat{Y}_i) & \text{otherwise} \end{cases} \quad (4)$$

where $\hat{Y}_i$ and $Y_i$ refer to the predicted and the ground truth value respectively, and α and β are hyperparameters for balancing difficult and easy samples. The values of α and β are empirically set to 2 and 4.

**Box vertices.** We use the four vectors from box vertices to the center point to describe the OBB, containing top-left $a$, top-right $b$, bottom-right $c$ and bottom-left $d$, as shown in Fig. 5. We use the proposed method MDL to replace the common ln-norm losses due to MDL's property of scale-invariance and its high consistency with detection metric. However, like other eight-parameter methods, MDL still cannot solve the problem of boundary discontinuity due to the periodicity of angle. As for this problem, we are inspired by the modulated rotation loss introduced in RSDet [7]. RSDet first sorts the predicts points and then takes the minimum loss of sorted points themselves, the points moved forward and backward by one place. We extend this method by dropping the sorting step and taking the minimum loss of four losses to make MDL continuous at boundary cases:

$$L_b = min \begin{cases} MDL(B(a,b,c,d), B^*(a^*,b^*,c^*,d^*)) \\ MDL(B(b,c,d,a), B^*(a^*,b^*,c^*,d^*)) \\ MDL(B(c,d,a,b), B^*(a^*,b^*,c^*,d^*)) \\ MDL(B(d,a,b,c), B^*(a^*,b^*,c^*,d^*)) \end{cases} \quad (4)$$

where $B(a,b,c,d)$ and $B^*(a^*,b^*,c^*,d^*)$ refer to the predicted and the ground truth box respectively, and $MDL(B, B^*)$ is according to Eq. 3.

**Offset.** As the image is downsampled by the factor of 4 in Centernet, there will be an accuracy damage when the feature map is remapped to the original image. Therefore, an additional offset is used to compensate for the damage. The ground truth is the difference between the target center point's coordinates downsampled by 4 and their integer values. Although offset loss is optional, for higher precision we choose to adopt this. We also use Mahalanobis distance to make the offset loss adapted to the size of the OBB. The loss between the predicted offset $O = (x, y)$ and the ground truth offset $O^* = (x^*, y^*)$ is calculated as below:

$$L_o = MD(O, O^*) = \sqrt{\left((x,y)-(x^*,y^*)\right)^T \Sigma^{-1}\left((x,y)-(x^*,y^*)\right)} \quad (5)$$

where Σ refers to the same covariance matrix as in **Box vertices**.

In summary, our overall loss function is designed as below:

$$L = \frac{1}{N}(L_h + L_b + L_o) \quad (6)$$

where $N$ indicates the number of ground truth objects.

## III. EXPERIMENTS

### A. Dataset

We use DOTA-v1.0 [9] as the dataset to validate the effectiveness of our method. The DOTA-v1.0 dataset contains 2806 aerial images and a total of 188282 instances in 15 categories: plane (PL), ship (SH), storage tank (ST), baseball diamond (BD), tennis court (TC), basketball court (BC), ground track field (GTF), harbor (HB), bridge (BR), large vehicle (LV), small vehicle (SV), helicopter (HC), roundabout (RA), soccer ball field (SBF) and swimming pool (SP). As most of the images are large in size, we crop the images into 600*600 patches with the gap of 100 at the scale 0.5 and 1.0. After cropping, we get a trainval set of 69,337 images and a test set of 35,777 images in total.

### B. Implementation and Testing Details

**Implementation Details.** The experiments are implemented using PyTorch. We use Adam [13] as our optimizer with an in-itial learning rate $1.25\times10^{-4}$. To make the network better converge to the optimal solution, we bind an exponentially decaying learning rate scheduler to the optimizer with the decay factor 0.96. For increasing the number and diversity of training samples, images are preprocessed by data augmentation including random cropping and random flipping during training. We use a batch size of 48 over 6 GeForce RTXTM 3090 GPUs to implement our experiments.

**Testing Details.** All the results for testing are derived from the 50 epoch model trained on DOTA-v1.0. From the output heatmaps, we adopt the top-500 points whose scores are more than 0.1 as the center points of objects. To obtain the coordinates of the predicted box, we first add the coordinates of the center point and the corresponding offset to get accurate center coordinates, then add the center coordinates and the box vertices' vectors, and finally multiply the added values by the downsampling factor 4. As we use multi-scale images (0.5 and 1.0) for testing, we apply Non-maximun-suppression (NMS) [14] to the output results with the threshold of 0.1 to get final merged results.

*C. Ablation Study*

**Ablation test of MDL forms**. As mentioned before, the covariance matrix Σ used in MDL can be calculated over the ground truth box vectors or the predicted box vectors. Therefore, we implement experiments to compare these two kinds of MDL forms. Table I shows the performances of MDL with covariance matrix using the target oriented box vectors (MDL-t) and using predicted vectors (MDL-p). Under the same condition, the mAP on DOTA-v1.0 of MDL-t is 74.33% while MDL-p achieves 76.16%. The fact that the mAP of MDL-p is 1.83% higher than that of MDL-t shows that using the covariance matrix calculated by the predicted box points can get better performance. The reason would probably be that using covariance matrix calculated by predicted points can further facilitate the regression of predicted values.

**Ablation study with smooth L1 loss.** To prove the effectiveness of our method, we compare MDL with smooth L1 loss, which is widely used in eight-parameter rotated object detection. Sharing the same architecture, implementation and testing details as MDL, we implement the experiment using smooth L1 loss for both box vertices and offset. We also adopt the minimum loss value for box vertices to avoid boundary discontinuity as MDL. As can be seen in Table I, the mAP on DOTA-v1.0 of smooth L1 loss is only 73.98%, which is lower than both MDL-t and MDL-p. In particular, MDL-p achieves 2.18% improvement over smooth L1 loss, demonstrating the superiority of our approach. Furthermore, Fig. 6 shows MDL is more stable during training than smooth L1 loss.

TABLE I
ABLATION TEST OF MDL FORMS AND ABLATION STUDY WITH SMOOTH L1 LOSS ON DOTA-v1.0.

| Method | mAP |
|---|---|
| smooth L1 loss | 73.98 |
| MDL-t (ours) | 74.33 (**+0.35**) |
| MDL-p (ours) | 76.16 (**+2.18**) |

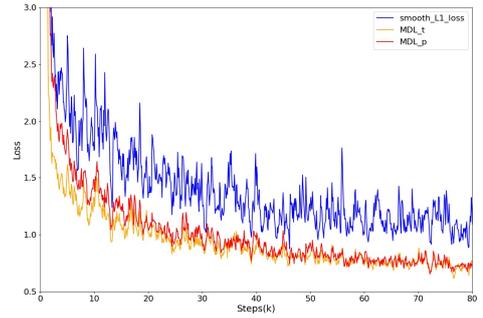

Fig. 6. Loss curves of smooth L1 loss, MDL-t and MDL-p during training.

*D. Further Comparison*

For further comparing the proposed method MDL with other methods, we choose both five-parameter and eight-parameter methods with different architectures, as shown in Table II. Most of these methods use smooth L1 loss as their loss functions, except for SCRDet [15] and RSDet [7], which introduces IoU-Smooth L1 loss based on smooth L1 loss and uses the modulated rotation loss based on L1 loss respectively. From Table II, we find that most results of eight-parameter methods are better than five-parameter methods, which implies that eight-parameter methods do solve the problem of measurement discontinuity in five-parameter methods to some extent. Among these methods, SCRDet gains over other five-parameter methods while RSDet and BBAVectors [1] achieve the top-2 performance in eight-parameter methods. SCRDet uses improved IoU-Smooth L1 loss to alleviate the loss upsurge at boundary cases and gains 72.61% in mAP. RSDet takes the minimum loss to make loss continuous at boundary, obtaining the accuracy of 74.1%. BBAVectors adds a parameter to predict the horizontality of the OBB and additionally use two parameters to predict width and height of its horizontal bounding box (HBB), which achieves 75.36%. These methods all manage to solve the boundary discontinuity problem, which indicates that boundary discontinuity hurts performance a lot and solutions should be taken in order to gain better performance.

By using Mahalanobis distance to calculate loss and taking the minimum loss value to solve the problem of boundary discontinuity, the proposed method MDL-p achieves 76.16% in mAP, exceeding all above methods. As most of these methods use smooth L1 loss as the loss function, the results of MDL-p show the superiority of MDL. Besides, MDL-p gains the best or second-best results in most object categories. In particular, MDL-p outperforms other methods in terms of Harbor (73.39%), Large Vehicle (81.25%) and Helicopter (68.31%). For further inspection, the visualization results of MDL-p on DOTA-v1.0 are illustrated in Fig. 7. Although DOTA-v1.0 consists of complex images with targets of different sizes and types, our proposed method can still achieve good performance.

IV. CONCLUSION

In this paper, we propose a new loss function MDL for eight-parameter rotated object detection. The proposed method MDL calculates the loss between predicted and target box vertices' vectors via scale-invariant Mahalanobis distance, which can

TABLE II

COMPARISON WITH DIFFERENT METHODS ON DOTA-v1.0. THE **RED** AND **BLUE** REFER TO THE TOP TWO PERFORMANCES.

| | PL | SH | ST | BD | TC | BC | GTF | HB | BR | LV | SV | HC | RA | SBF | SP | mAP |
|---|---|---|---|---|---|---|---|---|---|---|---|---|---|---|---|---|
| **Five-parameter methods** | | | | | | | | | | | | | | | | |
| FR-O [9] | 79.42 | 37.16 | 59.28 | 77.13 | 89.41 | 69.64 | 64.04 | 47.89 | 17.7 | 38.02 | 35.3 | 46.3 | 52.19 | 50.3 | 47.4 | 54.13 |
| ROITrans [16] | 88.64 | 66.57 | 76.75 | 78.52 | 90.5 | 79.46 | **75.92** | 62.54 | 43.44 | 62.97 | 68.81 | 55.56 | 56.73 | 59.04 | 61.29 | 69.56 |
| CAD-Net [17] | 87.80 | 76.60 | 73.30 | 82.40 | **90.90** | 79.20 | 73.50 | 62.00 | 49.40 | 63.50 | 71.10 | 62.20 | 60.90 | 48.40 | 67.00 | 69.90 |
| SCRDet [15] | **89.98** | 72.41 | **86.86** | 80.65 | 90.85 | **87.94** | 68.36 | 66.25 | 52.09 | 60.32 | 68.36 | 65.21 | **66.68** | **65.02** | 68.24 | 72.61 |
| **Eight-parameter methods** | | | | | | | | | | | | | | | | |
| ICN [2] | 81.36 | 69.98 | 78.20 | 74.30 | 90.76 | 79.06 | 70.32 | 67.02 | 47.70 | 67.82 | 64.89 | 50.23 | 62.90 | 53.64 | 64.17 | 68.16 |
| O$^2$-DNet [18] | 89.30 | 78.70 | 82.90 | 83.30 | **90.90** | 79.90 | 72.10 | 64.60 | 50.10 | 75.60 | 71.10 | **65.70** | 60.00 | 60.20 | 68.90 | 72.80 |
| RSDet [7] | **90.1** | 73.6 | 84.7 | 82.0 | **91.2** | 87.1 | 68.5 | 66.1 | **53.8** | 78.7 | 70.2 | 63.7 | **68.2** | 64.3 | 69.3 | 74.1 |
| BBAVectors[1] | 88.63 | **88.06** | 86.39 | **84.06** | 90.87 | 87.23 | 74.08 | **67.10** | 52.13 | **80.40** | **78.26** | 63.96 | 65.62 | 56.11 | **72.08** | 75.36 |
| **MDL-p(ours)** | 88.75 | **87.58** | **86.94** | **84.96** | 90.84 | **87.59** | 70.96 | **73.39** | **53.20** | **81.25** | 76.46 | **68.31** | 63.47 | 57.66 | **71.07** | **76.16** |

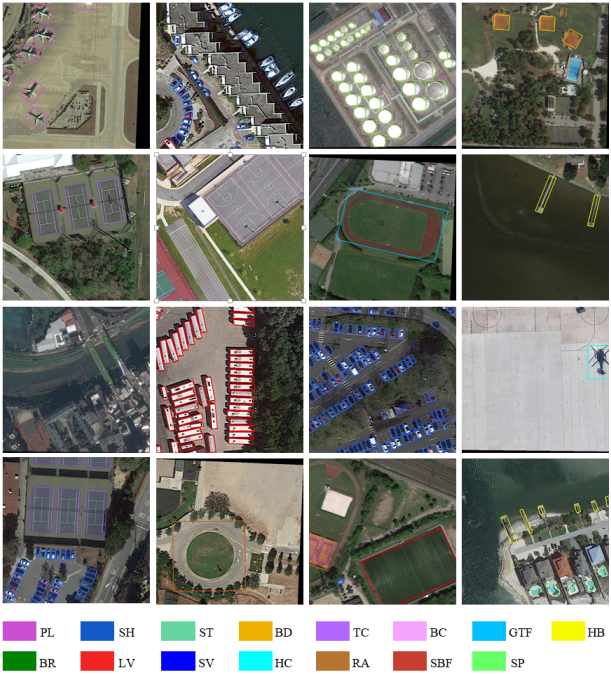

Fig. 7. Visualization results of MDL-p on DOTA-v1.0.

thus alleviate the problems of inconsistency with the detection metric SkewIoU and training instability when using $l_n$-norm losses. The experimental results on DOTA-v1.0 confirm the effectiveness of MDL. Besides, the experiments also show that in rotated object detection MDL performs better than smooth L1 loss, which is widely used in eight-parameter systems. However, like all other eight-parameter methods, MDL still faces the problem of boundary discontinuity, which we manage to solve by taking the minimum loss value in this paper. Therefore, in the future we would like to explore a better way to fundamentally solve boundary discontinuity for all eight-parameter methods. Furthermore, we would like to apply MDL to quadrilateral or polygon object detection as MDL can also be used on point sets.

REFERENCES

[1] J. Yi, P. Wu, B. Liu, Q. Huang, H. Qu, and D. Metaxas, "Oriented object detection in aerial images with box boundary-aware vectors," in *Proceedings of the IEEE/CVF Winter Conference on Applications of Computer Vision*, 2021, pp. 2150–2159.
[2] S. M. Azimi, E. Vig, R. Bahmanyar, M. Körner, and P. Reinartz, "Towards multi-class object detection in unconstrained remote sensing imagery," in *Asian Conference on Computer Vision*, 2018, pp. 150–165.
[3] J. Yu, Y. Jiang, Z. Wang, Z. Cao, and T. Huang, "Unitbox: An advanced object detection network," in *Proceedings of the 24th ACM international conference on Multimedia*, 2016, pp. 516–520.
[4] H. Rezatofighi, N. Tsoi, J. Gwak, A. Sadeghian, I. Reid, and S. Savarese, "Generalized intersection over union: A metric and a loss for bounding box regression," in *Proceedings of the IEEE/CVF conference on computer vision and pattern recognition*, 2019, pp. 658–666.
[5] Z. Zheng, P. Wang, W. Liu, J. Li, R. Ye, and D. Ren, "Distance-IoU loss: Faster and better learning for bounding box regression," in *Proceedings of the AAAI Conference on Artificial Intelligence*, 2020, vol. 34, no. 07, pp. 12993–13000.
[6] R. Girshick, "Fast r-cnn," in *Proceedings of the IEEE international conference on computer vision*, 2015, pp. 1440–1448.
[7] W. Qian, X. Yang, S. Peng, Y. Guo, and J. Yan, "Learning modulated loss for rotated object detection," *ArXiv Prepr. ArXiv191108299*, 2019.
[8] R. De Maesschalck, D. Jouan-Rimbaud, and D. L. Massart, "The mahalanobis distance," *Chemom. Intell. Lab. Syst.*, vol. 50, no. 1, pp. 1–18, 2000.
[9] G.-S. Xia *et al.*, "DOTA: A large-scale dataset for object detection in aerial images," in *Proceedings of the IEEE conference on computer vision and pattern recognition*, 2018, pp. 3974–3983.
[10] K. Duan, S. Bai, L. Xie, H. Qi, Q. Huang, and Q. Tian, "Centernet: Keypoint triplets for object detection," in *Proceedings of the IEEE/CVF international conference on computer vision*, 2019, pp. 6569–6578.
[11] X. Zhou, D. Wang, and P. Krähenbühl, "Objects as points," *ArXiv Prepr. ArXiv190407850*, 2019.
[12] H. Law and J. Deng, "Cornernet: Detecting objects as paired keypoints," in *Proceedings of the European conference on computer vision (ECCV)*, 2018, pp. 734–750.
[13] D. P. Kingma and J. Ba, "Adam: A method for stochastic optimization," *ArXiv Prepr. ArXiv14126980*, 2014.
[14] A. Neubeck and L. Van Gool, "Efficient non-maximum suppression," in *18th International Conference on Pattern Recognition (ICPR'06)*, 2006, vol. 3, pp. 850–855.
[15] X. Yang *et al.*, "Scrdet: Towards more robust detection for small, cluttered and rotated objects," in *Proceedings of the IEEE/CVF International Conference on Computer Vision*, 2019, pp. 8232–8241.
[16] J. Ding, N. Xue, Y. Long, G.-S. Xia, and Q. Lu, "Learning roi transformer for oriented object detection in aerial images," in *Proceedings of the IEEE/CVF Conference on Computer Vision and Pattern Recognition*, 2019, pp. 2849–2858.
[17] G. Zhang, S. Lu, and W. Zhang, "CAD-Net: A context-aware detection network for objects in remote sensing imagery," *IEEE Trans. Geosci. Remote Sens.*, vol. 57, no. 12, pp. 10015–10024, 2019.
[18] H. Wei, Y. Zhang, Z. Chang, H. Li, H. Wang, and X. Sun, "Oriented objects as pairs of middle lines," *ISPRS J. Photogramm. Remote Sens.*, vol. 169, pp. 268–279, 2020.